\documentclass[11pt]{article}

\usepackage[]{acl} 
\usepackage{times}
\usepackage{latexsym}
\usepackage{booktabs}
\usepackage{longtable}
\usepackage{amsmath}
\usepackage{graphicx}
\usepackage[T1]{fontenc}
\usepackage[utf8]{inputenc}
\usepackage{microtype}
\usepackage{inconsolata}
\usepackage{placeins}
\usepackage[table]{xcolor}

\definecolor{xlamcolor}{RGB}{207,226,243} 
\definecolor{tacolor}{RGB}{252,229,205}   
\definecolor{glcolor}{RGB}{217,234,211}   

\title{SFT or RL for Tool-Calling Agents? \\ A Controlled Study Across Data, Method, and Scale\\
\texorpdfstring{%
  \raisebox{-0.2\height}{\includegraphics[height=7mm]{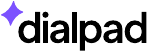}}%
}{}}

\author{Md Tahmid Rahman Laskar, Xue-Yong Fu, Shashi Bhushan TN\\\{\texttt{tahmid.rahman,xue-yong,sbhushan}\}\texttt{@dialpad.com}\\{Dialpad Inc.}}

\begin{document}
\maketitle

\begin{abstract}
Limited controlled evidence exists on how training data, adaptation method, and model scale jointly affect tool-calling performance in language-model agents. We evaluate supervised fine-tuning (SFT) with LoRA, reinforcement learning (RL) via Group Relative Policy Optimization (GRPO), and SFT followed by GRPO across six Qwen3 models from 0.6B to 32B parameters, covering both in-distribution performance and cross-dataset transfer. SFT with LoRA is the strongest in-distribution method throughout the 0.6B--32B range and best in 15 out of 18 experimental settings. 
On cross-dataset transfer, the methods are closer: GRPO wins 29 out of 54 settings where training and test datasets differ, but its margin over SFT averages under one point, and SFT$\rightarrow$GRPO is rarely strongest at either comparison. Dataset mixing gives consistently strong transfer while staying close to specialized in-distribution training, regardless of method. Additional analysis further confirms that LoRA outperforms full-parameter fine-tuning, 
demonstrating that 
LoRA better preserves pretrained agentic behavior.
\end{abstract}

\section{Introduction}

    Tool use is a core capability of language-model agents, enabling them to retrieve external information, interact with environments, and execute planned actions \citep{yao2022react,schick2023toolformer,li2023apibank}. Reliable tool use requires selecting the appropriate function and generating a valid call with correct arguments; failures at this can prevent the completion of tasks that require external actions \citep{li2023apibank,patil2024bfcl}. Despite its importance, practical guidance for training tool-calling models remains limited because existing studies often do not clearly separate the effects of training data, post-training strategy, and model scale. Moreover, constructing a sufficiently large and diverse tool-calling dataset from scratch is costly, making existing public corpora a practical starting point for model adaptation \cite{laskar2026text}. Practitioners must therefore jointly determine (i)~which tool-calling corpus to use, (ii)~whether to apply supervised fine-tuning (SFT), reinforcement learning (RL), or SFT followed by RL, and (iii)~whether the selected strategy remains effective at the intended deployment scale. 
    
  To address this gap, we conduct a controlled comparison of three post-training strategies across multiple tool-calling training corpora and model scales: SFT, RL applied directly to the instruction-tuned model, and SFT followed by RL. We use LoRA \citep{hu2022lora} for SFT and implement the RL-based strategy using Group Relative Policy Optimization (GRPO) \citep{shao2024deepseekmath}. Our experiments use three public tool-calling datasets: xLAM/APIGen 60k \citep{liu2024apigen}, ToolACE \citep{liu2025toolace}, and Glaive-FC-v2 \citep{glaive2023}, plus a uniform mixture of the three. We evaluate these strategies across six Qwen3 model sizes \citep{yang2025qwen3}, ranging from 0.6B to 32B. 

  To ensure a controlled and fair comparison, we use identical prompt formats across all datasets and also remove potential duplicates to mitigate the risk of data contamination between training and test splits \cite{ravaut2025a}. For SFT$\rightarrow$GRPO, each training set is divided equally and without replacement between the two stages. We evaluate every model on held-out test sets from all three datasets, allowing us to also investigate cross-dataset transfer. This design lets us investigate the following research questions (RQ): (RQ1) how SFT and GRPO compare across scale, in-distribution and under transfer, and (RQ2) how much training-corpus choice affects performance and transfer, independent of method.

\noindent \textbf{Findings.}
Four findings emerge. First, SFT with LoRA provides the most reliable in-distribution performance across the full 0.6B--32B scale range and improves over zero-shot in all 18 settings. Across these comparisons, SFT with LoRA achieves the highest accuracy in 15 settings, while GRPO performs best in only 2 and SFT$\rightarrow$GRPO in only 1. Second, GRPO offers a modest advantage in cross-dataset transfer. Among the 54 settings in which the training and test datasets differ, GRPO achieves the highest accuracy in 29, SFT in 19, and SFT$\rightarrow$GRPO in 6. However, GRPO's average improvement over SFT is less than one point. Third, combining SFT and GRPO provides no consistent benefit under a matched data budget and is rarely the strongest strategy for either in-distribution performance or cross-dataset transfer. Finally, training-corpus choice matters as much as the adaptation method, changing in-distribution accuracy by up to 55 points. The uniform mixture provides the strongest overall deployment default, remaining close to the best in-distribution specialist while achieving the best transfer performance. We additionally compare LoRA with full-parameter fine-tuning (full SFT) and observe that LoRA consistently outperforms full SFT, suggesting that it better preserves pretrained tool-use capabilities. 

\begin{table}[t]
\centering\small
\setlength{\tabcolsep}{3pt}
\begin{tabular}{l rrr}
\toprule
 & xLAM & ToolACE & Glaive \\
\midrule
Final train examples & 56,443 & 9,715 & 46,204 \\
Near duplicates removed (\%) & 3.5 & 0.9 & 58.5 \\
Multi-turn (\%) & 0.0 & 4.5 & 92.6 \\
Parallel calls (\%) & 54.2 & 17.2 & 0.0 \\
No-call / refusal (\%) & 0.0 & 56.1 & 73.3 \\
Avg.\ calls per example & 1.69 & 0.79 & 0.34 \\
\bottomrule
\end{tabular}
\caption{\small{Training-data statistics after near-duplicate removal.}}
\label{tab:data}
\end{table}

\section{Related Work}

\noindent \textbf{Tool use in agent language models.}
Tool calling enables language-model agents to interact with external systems and execute actions produced through planning, memory, and reasoning~\cite{yao2024tau}. Reliable execution requires the model to select the appropriate tool and generate a valid call with correct arguments \cite{laskar2026text}. Several datasets support training this capability but differ substantially in their construction and interaction patterns. APIGen/xLAM~\cite{liu2024apigen} provides 60k verified single-turn examples covering 4,211 APIs; ToolACE~\cite{liu2025toolace} uses multi-agent, self-evolving synthesis to generate conversations involving multi-turn interactions, parallel calls, and tool-use refusals across 26.5k APIs; and Glaive-FC-v2~\cite{glaive2023} contains 113k conversational examples combining ordinary dialogue with function calls. Despite these differences, limited evidence is available on how the choice of training corpus affects in-distribution performance and transfer to other tool-calling datasets.

\noindent \textbf{RL for tool use.}
Recent studies show that reinforcement learning can improve tool-calling performance. ToolRL~\cite{qian2025toolrl} applies GRPO directly to instruction-tuned models using rewards that separately assess output format, tool selection, and argument correctness. Across Qwen2.5 models at 1.5B, 3B, and 7B parameters, it reports that this RL-from-instruct approach generally outperforms the corresponding base and SFT models, although the gains vary across models and benchmarks. Nemotron-Tool-N1~\cite{zhang2025nemotron} similarly applies GRPO to Qwen2.5 models, using an R1-style binary reward~\cite{guo2025deepseekr1} that requires both a valid reasoning format and a correct tool call. Its 7B and 14B models outperform several existing SFT baselines on BFCL and API-Bank. However, neither study systematically compares RL against a matched parameter-efficient SFT baseline across a broad range of model scales and training corpora. Our results provide this comparison from 0.6B to 32B parameters: SFT with LoRA is consistently stronger in-distribution, whereas GRPO provides only a modest advantage under cross-dataset transfer.


\definecolor{xlamcolor}{RGB}{207,226,243}      
\definecolor{tacolor}{RGB}{252,229,205}        
\definecolor{glcolor}{RGB}{217,234,211}        
\definecolor{belowzerocolor}{RGB}{244,204,204} 

\begin{table*}[t]
\centering
\small
\begin{tabular}{ll ccc ccc ccc}
\toprule
 & & \multicolumn{3}{c}{SFT}
 & \multicolumn{3}{c}{GRPO}
 & \multicolumn{3}{c}{SFT$\rightarrow$GRPO} \\
\cmidrule(lr){3-5}
\cmidrule(lr){6-8}
\cmidrule(lr){9-11}
Model & Train data
& xLAM & TA & GL
& xLAM & TA & GL
& xLAM & TA & GL \\
\midrule

Qwen3-0.6B & \emph{none (0-shot)}
& 66.0 & 19.9 & 42.0
& \multicolumn{3}{c}{---}
& \multicolumn{3}{c}{---} \\

 & xLAM
& \cellcolor{xlamcolor}\textbf{85.6} & 22.3 & 40.5
& 79.2 & \textbf{24.6} & \textbf{42.0}
& 81.1 & 22.7 & 40.8 \\

 & ToolACE
& \textbf{65.0} & \cellcolor{tacolor}\textbf{26.0} & 42.9
& 46.8 & 20.6 & \textbf{43.9}
& 41.0 & 21.5 & 43.4 \\

 & Glaive
& 45.5 & 17.1 & 44.9
& \textbf{70.9} & \textbf{23.1} & \textbf{45.1}
& 54.4 & 19.3 & 44.2 \\

 & Mix
& 77.2 & \textbf{24.8} & \cellcolor{glcolor}\textbf{45.4}
& \textbf{78.7} & 20.9 & 44.8
& 78.6 & 22.9 & 44.4 \\

\midrule
Qwen3-1.7B & \emph{none (0-shot)}
& 68.1 & 22.7 & 42.5
& \multicolumn{3}{c}{---}
& \multicolumn{3}{c}{---} \\

 & xLAM
& \cellcolor{xlamcolor}\textbf{87.2} & 23.9 & 40.2
& 81.4 & \textbf{27.1} & \textbf{44.6}
& 82.6 & 23.8 & 40.2 \\

 & ToolACE
& \textbf{73.8} & \cellcolor{tacolor}\textbf{29.6} & \textbf{44.6}
& 73.5 & 26.0 & 43.2
& 44.4 & 25.5 & 43.2 \\

 & Glaive
& 43.5 & 19.2 & \cellcolor{glcolor}\textbf{45.9}
& \textbf{72.8} & \textbf{26.7} & 45.6
& 37.9 & 17.8 & \cellcolor{belowzerocolor}40.8 \\

 & Mix
& \textbf{81.1} & 28.8 & 45.3
& 80.1 & \textbf{29.1} & \textbf{45.8}
& 80.2 & 27.1 & 45.8 \\

\midrule
Qwen3-4B & \emph{none (0-shot)}
& 76.6 & 28.1 & 45.3
& \multicolumn{3}{c}{---}
& \multicolumn{3}{c}{---} \\

 & xLAM
& \cellcolor{xlamcolor}\textbf{88.0} & 26.1 & 43.6
& 83.3 & \textbf{28.2} & \textbf{45.3}
& 84.2 & 26.9 & 43.7 \\

 & ToolACE
& \textbf{75.5} & \textbf{30.4} & \textbf{44.4}
& 41.4 & \cellcolor{belowzerocolor}25.5 & 43.9
& 58.6 & 29.4 & 44.2 \\

 & Glaive
& 68.4 & 22.5 & \cellcolor{glcolor}\textbf{47.6}
& \textbf{79.2} & \textbf{29.0} & 46.1
& 76.0 & 27.8 & \cellcolor{belowzerocolor}42.4 \\

 & Mix
& \textbf{82.5} & 30.5 & \textbf{46.3}
& 81.9 & 30.5 & 46.1
& 82.5 & \cellcolor{tacolor}\textbf{31.0} & 46.3 \\

\midrule
Qwen3-8B & \emph{none (0-shot)}
& 75.3 & 27.6 & 43.9
& \multicolumn{3}{c}{---}
& \multicolumn{3}{c}{---} \\

 & xLAM
& \cellcolor{xlamcolor}\textbf{87.7} & 26.8 & \textbf{44.6}
& 84.4 & \textbf{29.5} & \textbf{44.6}
& 86.2 & 28.2 & 44.4 \\

 & ToolACE
& 75.9 & \cellcolor{tacolor}\textbf{31.5} & 44.4
& \textbf{77.3} & 30.9 & 43.9
& 76.1 & 28.8 & \textbf{44.9} \\

 & Glaive
& 74.4 & 25.6 & \cellcolor{glcolor}\textbf{47.5}
& \textbf{77.7} & \textbf{29.0} & 46.4
& 69.2 & 27.3 & 46.6 \\

 & Mix
& \textbf{84.6} & \cellcolor{tacolor}\textbf{31.5} & 45.4
& 82.0 & 31.4 & 46.1
& 84.1 & 30.9 & \textbf{46.3} \\

\midrule
Qwen3-14B & \emph{none (0-shot)}
& 77.8 & 29.0 & 46.1
& \multicolumn{3}{c}{---}
& \multicolumn{3}{c}{---} \\

 & xLAM
& \cellcolor{xlamcolor}\textbf{89.7} & 27.2 & 42.4
& 85.1 & \textbf{29.8} & \textbf{46.1}
& 86.2 & 28.2 & 44.4 \\

 & ToolACE
& \textbf{79.9} & \textbf{32.4} & 42.6
& 79.3 & 30.1 & \textbf{45.9}
& 77.6 & 31.7 & 44.9 \\

 & Glaive
& 76.7 & 26.7 & 47.1
& \textbf{79.0} & \textbf{29.7} & 47.3
& 75.8 & 28.5 & \cellcolor{glcolor}\textbf{48.3} \\

 & Mix
& \textbf{83.9} & 32.3 & 45.9
& 82.3 & 32.0 & \textbf{46.8}
& 81.7 & \cellcolor{tacolor}\textbf{32.5} & 46.6 \\

\midrule
Qwen3-32B & \emph{none (0-shot)}
& 78.3 & 28.5 & 44.8
& \multicolumn{3}{c}{---}
& \multicolumn{3}{c}{---} \\

 & xLAM
& \cellcolor{xlamcolor}\textbf{88.7} & 28.5 & 43.4
& 85.8 & \textbf{30.0} & 43.7
& 86.0 & 29.4 & \textbf{43.9} \\

 & ToolACE
& \textbf{81.5} & \textbf{32.9} & 43.6
& 57.0 & 28.7 & \textbf{44.9}
& 74.5 & 30.1 & 43.0 \\

 & Glaive
& \textbf{80.5} & 27.9 & 46.9
& 79.3 & \textbf{30.3} & \cellcolor{glcolor}\textbf{47.3}
& 77.4 & 25.8 & \cellcolor{belowzerocolor}43.6 \\

 & Mix
& 85.8 & \cellcolor{tacolor}\textbf{33.1} & \textbf{46.8}
& 84.3 & 31.9 & \textbf{46.8}
& \textbf{86.1} & 32.0 & 40.3 \\

\bottomrule
\end{tabular}

\caption{\small{Exact-match accuracy (\%) on the three held-out test
splits (xLAM, ToolACE=TA, and Glaive=GL) for every model
$\times$ training-dataset $\times$ method combination, alongside
zero-shot baselines. 
Bold indicates the best method within each training-data setting.
Within each model scale, blue, orange, and green indicate the best
result on xLAM, ToolACE, and Glaive, respectively. Red indicates an
in-distribution result below the corresponding zero-shot baseline.
Tied best results are all highlighted.}}
\label{tab:main}
\end{table*}
\section{Experimental Setup}

\subsection{Data}
We use the following three datasets in this paper: xLAM~\citep{liu2024apigen}, ToolACE~\citep{liu2025toolace}, Glaive-FC-v2~\citep{glaive2023}. We convert all datasets to a common format containing tool definitions, messages, and reference calls. To prevent overlap between the training, validation, and test sets, we first remove near-duplicate examples from each corpus based on normalized user queries and tool names and only then create the splits. This removes 65{,}256 examples from Glaive, more than half of the original corpus. Using a fixed seed, we reserve 1{,}000 examples for testing and 500 for validation from each source; Table~\ref{tab:data} summarizes the resulting data.


We also use a dataset mixture that contains a maximum of randomly sampled 20{,}000 training examples from each source (if any dataset has fewer than 20,000 samples, we use all instances in that dataset). For SFT$\rightarrow$GRPO, we divide each training set equally without overlap: one half is used for SFT and the other for GRPO. Each example therefore appears in only one training stage, keeping the total amount of training data consistent across methods. We use the same system prompt and tool format for training and evaluation, ensuring that performance differences arise from the training methods rather than prompt-format differences. Appendix~\ref{app:prompt} provides the prompt and serialization format.
\subsection{Training}

\noindent \textbf{SFT.}
We apply LoRA ($r{=}32$, $\alpha{=}64$) at every model scale from 0.6B to 32B parameters, motivated by evidence that LoRA better preserves pretrained capabilities than full-parameter fine-tuning~\citep{biderman2024lora}. We implement using LLaMA-Factory~\citep{zheng2024llamafactory} and train for a maximum of two epochs using a cosine learning-rate schedule, bf16 precision, and loss on assistant turns only. See Appendix~\ref{app:sft-hparams} for the detailed hyperparameters.

\noindent \textbf{GRPO.}
We use a group size of 8, a KL coefficient of 0.001, a training batch size of 256, a mini-batch size of 64, and two epochs. We implement GRPO using verl~\citep{sheng2024hybridflow}, with vLLM~\citep{kwon2023vllm} generating the training responses. Appendix~\ref{app:grpo-hparams} provides the complete training and rollout settings. The reward combines output-format validity, tool-name correctness, and argument correctness with weights of 0.2, 0.3, and 0.5, respectively. It provides partial credit for correct arguments, penalizes additional arguments, and rewards the model for correctly predicting when no tool should be called. The complete reward definition is provided in Appendix~\ref{app:reward}.

\noindent \textbf{SFT$\rightarrow$GRPO.} This strategy performs cold-start SFT on one half of the training data (with the LoRA adapter merged into the base weights) and then applies GRPO to the disjoint remaining half.

\noindent All training jobs were run on Google Kubernetes Engine\footnote{\url{https://cloud.google.com/kubernetes-engine}} (GKE) on a single node with 8$\times$H200-141GB GPUs. Evaluation used data-parallel vLLM decoding at temperature 0 (Appendix~\ref{app:eval-hparams}) in a machine with 8$\times$A100-80GB  GPUs.

\subsection{Evaluation and metrics}

Each model is evaluated on the held-out test split of every dataset. We report exact match, defined as the proportion of examples for which all gold calls are reproduced \cite{laskar2025improving}. 
We write a parsing script to measure tool call correctness by comparing the tool call in the LLM-generated response with the gold tool call \cite{laskar2023systematic,laskar2024systematic,laskar2024query,laskar2025judging}. We use the same scoring function to compute these metrics and to compute the GRPO reward, ensuring exact parity between the training signal and the evaluation metric. We also normalize values before comparison (e.g., treating numeric strings as equivalent to numbers, and case-insensitive comparisons) so that equivalent representations are not penalized.

\section{Results}

Table~\ref{tab:main} summarizes our experimental results.

\subsection{RQ1: How does SFT compare to GRPO across scale?}
\label{sec:rq1}

In-distribution SFT improves over the zero-shot baseline in all 18 settings, with gains of up to 20 percentage points at 0.6B (xLAM: 66.0\%$\to$85.6\%). It is also the strongest in-distribution method overall, achieving the best result in 15 of 18 settings, compared with 2 for GRPO and 1 for SFT$\rightarrow$GRPO. GRPO and SFT$\rightarrow$GRPO fall below zero-shot in 1 and 3 settings, respectively, whereas SFT never does (Table~\ref{tab:main}). The methods are more competitive under cross-dataset transfer. Among the 54 settings in which the training and test datasets differ, GRPO achieves the best result in 29, SFT in 19, and SFT$\rightarrow$GRPO in 6. However, GRPO's average improvement over SFT is less than one point.

We additionally compare LoRA with full-parameter fine-tuning at 4B parameters, a scale at which practitioners may deploy agentic models in real-world settings~\cite{belcak2025small}. LoRA performs better on all three datasets: 88.0\% versus 80.4\% on xLAM, 30.4\% versus 22.8\% on ToolACE, and 47.6\% versus 44.1\% on Glaive. LoRA is also the only approach that never falls below zero-shot at this scale, consistent with evidence that it better preserves pretrained capabilities than full-parameter training~\citep{biderman2024lora}.

\subsection{RQ2: Dataset effects and transfer}

\begin{figure}[!t]
\centering
\includegraphics[width=0.9\columnwidth]{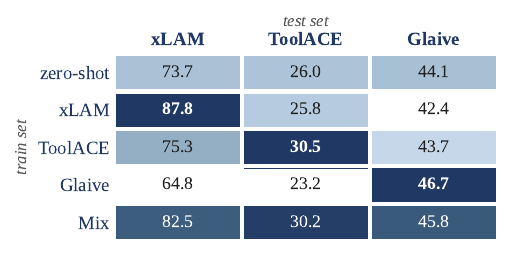}
\caption{\small{Cross-dataset transfer (exact match, SFT with LoRA, averaged over six Qwen3 scales).}}
\label{fig:transfer}
\end{figure}

The choice of training dataset has a largely consistent effect across model scales and influences performance as much as the training method (Figure~\ref{fig:transfer}, averaged across scales for SFT with LoRA). Models trained on ToolACE and the mixture transfer well to xLAM, achieving 75.3\% and 82.5\%, respectively, compared with a zero-shot average of 73.7\%. In contrast, Glaive-trained models achieve only 64.8\% on xLAM, below the zero-shot baseline. This suggests that Glaive's narrower, largely single-call format may not transfer well to other tool-calling settings. ToolACE is the most difficult target: even models trained directly on ToolACE achieve only 30.5\%, while models trained on the other individual datasets achieve between 23.2\% and 30.2\%. The uniform mixture provides the strongest overall deployment choice, remaining within 1--5 percentage points of the best dataset-specific model on every target while avoiding the poor Glaive-to-xLAM transfer. This pattern is consistent across all three training methods.

\section{Conclusion}

In this paper, our extensive experiments demonstrate that SFT with LoRA is the strongest in-distribution post-training strategy for tool-calling agents across the full 0.6B--32B range and never underperforms the zero-shot baseline. GRPO retains a modest advantage on cross-dataset transfer, which is most relevant when an agent operates outside its training distribution, and SFT$\rightarrow$GRPO is rarely the strongest strategy under either comparison. Dataset choice matters independently of method, and a seeded uniform mixture is the strongest default across scales. We will release our experimental resources to support further study here: \url{https://github.com/talkiq/dialpad-ai-research}.

\section*{Limitations}

One limitation is that we do not report a compute-normalized comparison since GRPO uses substantially more compute per training example than SFT. Our comparison against full-parameter fine-tuning is conducted at a single scale (4B parameters); whether LoRA's advantage over full fine-tuning holds at other scales is left to future work. Finally, all scale points in this study are drawn from a single model family, Qwen3, and replication in a second family would strengthen the generalizability. 

\section*{Ethics Statement}

All training data are public corpora used under their licenses (CC-BY-4.0, Apache-2.0). The study trains models to produce structured API calls; it involves no personal data or sensitive attributes. 

\bibliography{custom}

\appendix

\section{Reproducibility Details}
\label{app:repro}

\subsection{Prompt format}
\label{app:prompt}

The same Python module renders the system prompt for SFT training targets, GRPO rollout prompts, and evaluation prompts (verbatim, no paraphrasing across arms):

\begin{quote}
\small\ttfamily
You are a function-calling assistant. You may call one or more tools to answer the user. Available tools:\\
\{tools\}\\
For each tool call, output <tool\_call>\{"name": <function-name>, "arguments": <args-dict>\}</tool\_call>. If no available tool applies, answer directly without calling any tool.
\end{quote}

The placeholder \texttt{\{tools\}} denotes the JSON-serialized list of available tools for a given example, produced with \texttt{json.dumps} without indentation and with non-ASCII characters left unescaped. Assistant turns render each call as \texttt{<tool\_call>\{"name": ..., "arguments": ...\}</tool\_call>}, one call per line, matching the syntax used for scoring at evaluation time. External harnesses, namely BFCL and API-Bank, instead use each model's native chat template and tool-calling format; this is applied identically across all training conditions and is reported separately from our transfer matrix.

\subsection{SFT hyperparameters}
\label{app:sft-hparams}

We use LLaMA-Factory~\citep{zheng2024llamafactory} with LoRA at every model scale from 0.6B to 32B parameters, using rank $r{=}32$, $\alpha{=}64$, dropout 0.05, and applying the adapter to all linear layers. We optimize with AdamW at a learning rate of $1{\times}10^{-4}$, a cosine schedule with 5\% linear warmup, and two training epochs (one epoch for the SFT half of SFT$\rightarrow$GRPO, since that stage is a cold start on half the data preceding a subsequent GRPO stage). The effective batch size is 64, obtained from a per-device batch size of 4, gradient accumulation of 2, and 8 GPUs. We train in bf16 precision with gradient checkpointing enabled and a cutoff length of 4096 tokens, computing the loss on assistant turns only. The full-parameter fine-tuning baseline used for the additional LoRA-versus-full-fine-tuning comparison at 4B parameters in Section~\ref{sec:rq1} uses the same optimizer, schedule, and number of epochs, but a learning rate of $1{\times}10^{-5}$, a per-device batch size of 2, gradient accumulation of 4, and DeepSpeed ZeRO stage 2. The LoRA rank ablation, conducted at 8B parameters with $r \in \{16, 32, 64\}$, holds the ratio $\alpha/r{=}2$ fixed and varies only $r$ and $\alpha$, with all other hyperparameters unchanged.

\subsection{GRPO hyperparameters}
\label{app:grpo-hparams}

We use verl~\citep{sheng2024hybridflow} with the FSDP2 training strategy and a vLLM rollout backend~\citep{kwon2023vllm}. The advantage estimator is GRPO with a group size of 8. The actor is optimized with AdamW at a learning rate of $1{\times}10^{-6}$, using a PPO mini-batch size of 64, a micro-batch size of 8 per GPU for both the actor and the reference model, a training batch size of 256 prompts, and two epochs. We apply a KL-regularized loss with a KL coefficient of 0.001. During rollout, we use a vLLM tensor-parallel size of 1 (2 for the 32B model), a GPU memory utilization target of 0.6, a maximum prompt length of 3072 tokens, and a maximum response length of 1024 tokens, without parameter offloading for either the actor or the reference model. Each run uses 8 GPUs on a single node. SFT$\rightarrow$GRPO applies identical GRPO hyperparameters to the disjoint 50\% prompt split, initializing from the SFT stage's checkpoint; for models of 8B parameters and above, where SFT uses LoRA, the adapter is merged into the base weights before this second stage begins.

\subsection{Reward function}
\label{app:reward}

The reward is defined as $\max\!\big(0,\ 0.2\cdot\mathrm{fmt} + 0.3\cdot\mathrm{name\_frac} + 0.5\cdot\mathrm{arg\_frac} - 0.1\cdot\max(0, |\mathrm{pred}| - |\mathrm{gold}|)\big)$, where $\mathrm{fmt}$ equals 1 if the completion parses into a list of \texttt{\{name, arguments\}} calls and 0 otherwise, in which case the entire reward is 0. Gold and predicted calls are matched greedily and one-to-one by function name. The term $\mathrm{name\_frac}$ is the fraction of gold calls matched by name, and $\mathrm{arg\_frac}$ averages, over all matched pairs, the fraction of gold arguments whose predicted value equals the gold value after normalization (numeric strings are cast to numbers and comparisons are case-insensitive), multiplied by a penalty of $\max(0, 1-0.25\cdot\mathrm{extra})$ for predicted arguments absent from the gold set. Gold calls with no arguments receive an argument fraction of 1.0 if the prediction likewise has none, and 0.5 otherwise. When the gold label indicates that no tool call is appropriate, the reward is 1.0 for predicting no call and 0.0 for predicting any call. An ablation flag allows switching to a binary, Tool-N1-style reward, awarding 1.0 only when every gold call is matched exactly by name and by all arguments, and 0.0 otherwise; our main results use the decomposed reward described above. We use this same scoring function to compute our evaluation metrics, namely exact match, name accuracy, and argument score, which guarantees exact parity between the reward used in training and the metric used in evaluation.

\subsection{Evaluation}
\label{app:eval-hparams}

We perform offline inference with vLLM\footnote{\url{https://vllm.ai/}} at temperature 0 (greedy decoding), a maximum model length of 8192 tokens, and a maximum of 1024 new tokens. LoRA checkpoints are evaluated with the adapter loaded directly by vLLM (using \texttt{enable\_lora} with \texttt{max\_lora\_rank} set to the adapter's rank) applied to the corresponding base model, rather than merged in advance, so that evaluation exercises the exact artifact that would be deployed. Full-parameter checkpoints are loaded directly. We distribute evaluation in a data-parallel fashion across all available GPUs at a tensor-parallel size of 1 (2 for the 32B model, since a single GPU cannot hold its full-precision weights together with the key-value cache), replicating the model across the remaining GPUs.




\end{document}